\documentclass[letterpaper]{article} 
\usepackage{aaai2027}
\usepackage[hyphens]{url}  
\usepackage{graphicx} 
\graphicspath{ {./Figures/} }
\usepackage{natbib}  
\usepackage{caption} 
\usepackage{booktabs}
\usepackage{amsmath}
\usepackage{float}
\usepackage{placeins} 

\title{FTB Graph: Determining and Validating First-token Broadcasters and Language-Identity Head Circuits in Multilingual Language Models}

\author{
    Arjun Pillai\textsuperscript{\rm 1},
    Christian Hoang\textsuperscript{\rm 2},
    Anjelo Laroza\textsuperscript{\rm 3}
}
\affiliations{
  \textsuperscript{1}Irvington High School,   \textsuperscript{2}Gen4AIE, \textsuperscript{3}Independent Researcher
}
\date{}

\begin{document}

\maketitle

\begin{abstract}
Large language models operating in multilingual contexts must resolve target response languages early in generation, yet the exact causal circuitry governing first-token language identity decisions remains poorly mapped. We present an end-to-end structural circuit analysis across six model architectures spanning four families (GPT-2, BLOOM-560M, Pythia-1B/2.8B, and Qwen2.5-1.5B Base/Instruct). Utilizing Edge Attribution Patching (EAP) with FP16 active clamping, followed by exact activation patching verification bounded by a $2{,}000$ candidate edge search ceiling, we extract and map the directed acyclic graphs (DAGs) driving first-token language broadcasting. Our topological findings reveal three core mechanics: (1) The four standalone models exhibit a consistent deep-layer hub topology, where upstream residual signals aggregate into deep-layer broadcasting bottlenecks (e.g., Layer 21 in BLOOM) prior to output projection -- though this pattern is best supported for Pythia-2.8B and BLOOM-560M, since GPT-2 and Pythia-1B pass the necessity check with almost no out-of-graph heads remaining for comparison, and the Qwen2.5-1.5B pair (Layer 27 hub) inverts the check entirely rather than confirming it; (2) Parameter scaling (Pythia-1B to 2.8B) expands node participation (123 to 415 nodes) while maintaining a fixed edge budget ($\sim$860 edges), inducing high topological sparsity; and (3) Post-training alignment: the base and instruct DAGs for (Qwen2.5-1.5B) retain an $84.7\%$ Jaccard similarity ---including the primary Layer 27 broadcasting hub---demonstrating that first-token routing circuitry is fundamentally established during pre-training rather than synthesized during instruction tuning. Finally, by evaluating EAP attribution fidelity against exact patching deltas ($r \in [-0.23, 0.57]$), we show that linear gradient approximations frequently diverge from true causal interventions in low-precision regimes, proving that multi-stage exact patching verification is essential for reliable circuit discovery.
\end{abstract}

\section{Introduction}

Multilingual language models are routinely deployed in settings where the target response language must be inferred from context rather than stated explicitly --- from a code-switched prompt, from the language of a preceding turn, or from a single unambiguous word early in the input. This decision is made early: by the time a model emits its first token, it has already committed, implicitly, to a language. Yet little is known about \emph{where} in the network this commitment is made, \emph{how} it is computed, or whether the underlying circuitry is shared across model architectures, scales, and training regimes.

We study this question through the lens of mechanistic interpretability, focusing on attention heads whose activity at the first generated token is disproportionately predictive of the response's target language. We call such heads \emph{language identity heads} and refer to the broader phenomenon as \emph{first-token broadcasting}: the hypothesis that a small number of heads compute and propagate a language-identity signal early enough to constrain all subsequent generations.
\\
Our contributions are as follows:
\begin{itemize}
\item LIHA, a per-head attribution method for language-identity sensitivity at the first generated token, applied across six architectures (560M--1.5B parameters): four standalone models and a base-to-instruct pair, each with complete validation.
\item Functional circuit extraction via edge attribution patching, bounded by a $2{,}000$-edge candidate ceiling and FP16 active clamping, across all six models.
\item A base-to-instruct comparison within Qwen2.5-1.5B: both variants fail the necessity/irrelevance separation that holds for all four standalone models, while also showing the weakest EAP-to-exact-patching correlation of the six (Table~\ref{tab:eap-fidelity}) -- evidence we read as pipeline-reliability rather than an architecture-specific absence of broadcasting, since the same pattern appears in both variants.
\item An EAP-to-exact-patching correlation analysis across all six models (Table~\ref{tab:eap-fidelity}), as a validity check on the attribution stage given our use of fp16 throughout.
\end{itemize}

\section{Related Work}

\paragraph{Circuit discovery and Edge Attribution Patching}
Mechanistic circuit analysis seeks to identify a computational subgraph \citep{lindsey2025biology} that causally accounts for the model's behavior on targeted inputs and outputs. Automated Circuit Discovery (ACDC) performs this search by repeatedly applying activation-patching interventions and pruning edges \citep{bhaskar2024edge} that have a limited effect on a task-specific metric \citep{conmy2023acdc, meng2022locating}. Activation patching evaluates the model under an intervention, but exhaustive edge-level patching becomes expensive because candidates generally require separate patched executions.

Edge Attribution Patching (EAP) \citep{syed2024attribution} provides a scalable alternative for candidate screening. It uses a first-order Taylor approximation to estimate the effects of many edge interventions from two forward passes and one backward pass. This makes EAP suitable for searching the thousands of candidate edges in our evaluated models, but its scores remain linear approximations rather than exact causal effects. Similarly, this approach was implemented by \cite{marks2025sparse} to localize subgraphs of models via sparse autoencoders (SAE) features and gradient-based attribution methods across nodes and edges to isolate causally relevant features.

Critically, activation patching results are also sensitive to methodological choices beyond just the search algorithm: \citet{zhang2024towards} show that the choice of corruption method and evaluation metric can itself change which components a patching study identifies as important, independent of the search strategy used to find them -- a consideration that directly motivates the corruption scheme we fix in the Methodologies section below.

\paragraph{Language-selective components and representations}
Previous work has localized multilingual behavior in attention heads, feed-forward neurons, and intermediate representations. Studies of multilingual translation have found that many important attention heads are shared across language pairs, while some exhibit language-specific specialization \citep{kim2021multilingual, ferrando-costa-jussa-2024-similarity}. Shapley-based head pruning has also identified attention heads associated with multilingual interference \citep{held2023shapley}. At the neuron level, \citep{kojima2024multilingual} and \citep{tang2024languagespecific} identified sparse language-selective neurons and showed that intervening on a small proportion of neurons can alter or steer the generated language.

Representational analysis provides a complementary perspective: multilingual Llama-2 representations progress through input, intermediate concept, and output-language phases, with target-language realization becoming increasingly explicit in later layers \citep{wendler2024llamas, chang-etal-2022-geometry}. Our work differs by examining attention heads together with their directed connections and by focusing specifically on the language decision expressed at the first generated token.

\paragraph{First-token language mechanisms}
A growing body of interpretability work uses causal interventions on individual model components \citep{daiknowledge2022} -- attention heads, neurons, or subspaces -- to establish that language identity is represented in a distributed but partially localized fashion; \citet{resck2025explainability} survey this landscape of methods as applied to multilingual LLMs. As discussed above, \citet{tang2024languagespecific} showed that ablating a sparse set of language-specific neurons is sufficient to steer or disrupt a model's output language, establishing a causal link between individual components and language behavior. However, this style of intervention -- ablating a single component and observing its downstream effect on generation -- establishes \emph{that} a component affects language behavior without identifying the directed paths through which its signal propagates, or the specific point in generation at which the language decision is made.

The present work extends this line of inquiry from individual-component ablation to edge-defined computational graphs, and narrows the object of study to a specific decision point: rather than measuring aggregate effects on language behavior over a generated continuation, our pipeline measures target-language probability at the first generated token, tracing the directed edges responsible for that decision.

\paragraph{Circuit preservation under post-training}
Prior work has also examined whether fine-tuning creates new mechanisms or strengthens mechanisms already present in the base model. In an entity-tracking case study, base and fine-tuned models were found to rely primarily on the same underlying circuit, with fine-tuning improving the execution of the existing mechanism rather than replacing it \citep{prakash2024finetuning}.

\section{Methodologies}

\subsection{Pipeline Overview}

\begin{figure*}[t]
    \centering
    \includegraphics[width=\textwidth]{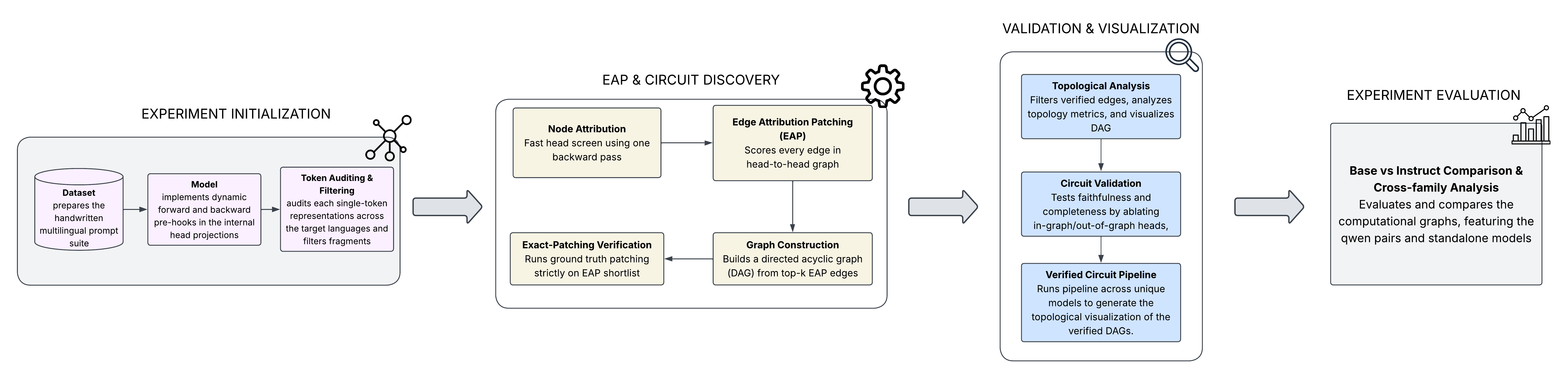}
    \caption{Overview of the proposed first-token broadcasting circuit-discovery
    pipeline. The workflow begins with dataset preparation, model
    instrumentation, and token-level auditing; applies node attribution and
    Edge Attribution Patching (EAP) to identify candidate circuit edges;
    verifies the shortlisted edges using exact activation patching; constructs
    the resulting directed acyclic graph; and validates the discovered circuit
    through topological analysis and causal ablation experiments.}
    \label{fig:ftb-pipeline}
\end{figure*}

Our circuit-discovery pipeline consists of six stages, applied identically to all six models:

\begin{enumerate}
\item \textbf{Dataset.} Consumption of the 21 hand-authored, neutral single-language prompt items (see Dataset, below).
\item \textbf{LIHA (Language Identity Head Attribution).} Scoring individual attention heads' sensitivity to target-language identity at the first generated token (see Language Identity Head Attribution, below).
\item \textbf{EAP (Edge Attribution Patching).} Fast-screening of the model to estimate edge contributions via a single backward pass, constrained by an FP16 active clamping mechanism and a strict ceiling of $2{,}000$ candidate edges (see Circuit Discovery via Edge Attribution Patching, below).
\item \textbf{Exact Activation Patching (Verification).} Rigorously pruning the $2{,}000$-edge candidate space by calculating exact-patching deltas and enforcing sign-agreement with the EAP scores (see Exact Activation Patching (Verification), below).
\item \textbf{DAG Extraction (Graph).} Rendering the final, pruned directed acyclic graph representing the verified first-token routing circuit for each model.
\item \textbf{Validation.} Executing boundary ablations to compute the completeness (necessity) drop and out-of-graph (irrelevance) drop (see Validation Protocol, below).
\end{enumerate}

\FloatBarrier

\subsection{Language Identity Head Attribution (LIHA)}

For each model, we score every attention head's contribution to target-language identity at the first generated token using activation patching between a target-language prompt and a corrupted foil in a different language. A head's LIHA score reflects the change in target-language log-probability attributable to that head in isolation. Associating this with the study of \citep{fierro2024how}, we draw a functional parallel to their work, where it was demonstrated that the last token position in input sequences is causally responsible for the change in the target output language, which mirrors our approach to isolate the causal drivers of language selection \citep{lindsey2025biology}. Positive scores indicate heads that, when active, push generation toward the correct target language; negative scores indicate heads whose activity is associated with the foil language. Table~\ref{tab:coverage} reports LIHA coverage (samples scored, total heads scored) and each model's single highest-magnitude head.

\subsection{Circuit Discovery via Edge Attribution Patching}

We apply EAP over the space of (source head $\rightarrow$ destination layer) edges, scoring each by its estimated contribution to the language-identity objective in one backward pass per sample. For an edge $e$ connecting upstream activation $z_i$ to downstream component $\psi_j$, the EAP score approximates the effect of patching $z_i$ from its target-language (clean) value $z_i^{\text{clean}}$ to its foil-language (corrupted) value $z_i^{\text{corr}}$ using a first-order Taylor expansion around the clean activation:
\begin{equation}
\phi_{\text{EAP}}(e) = \left(z_i^{\text{corr}} - z_i^{\text{clean}}\right)^{\top} \frac{\partial \mathcal{L}}{\partial z_i}\Bigg|_{z_i^{\text{clean}}},
\label{eq:eap}
\end{equation}
where $\mathcal{L}$ is the target-language log-probability loss. This requires only a single backward pass per sample rather than a separate forward pass per candidate edge, which allows EAP's tractability at the edge counts in Table~\ref{tab:coverage} (up to 15{,}872 for Pythia-2.8B) but also introduces the linear-approximation risk discussed in Limitations. The number of edges scored scales with model size: from 792 for GPT-2 (144 heads) to 15{,}872 for Pythia-2.8B (1024 heads) (Table~\ref{tab:coverage}).

\subsection{Exact Activation Patching (Verification)}

To maintain computational tractability and consistency across the six models, the verification framework enforces a strict upper ceiling of $2{,}000$ candidate edges per model prior to exact-patching verification; all six models reach this ceiling (Table~\ref{tab:coverage}). The final verified graph size, however, varies substantially by model after the sign-agreement filter is applied -- from 816 edges for GPT-2 to 1{,}016 for Qwen2.5-1.5B -- reflecting differences in how much of each model's candidate pool survives verification rather than differences in candidate pool size itself.

For each of the $2{,}000$ candidate edges, we compute an exact patching delta by actually substituting the corrupted activation into the clean forward pass and measuring the resulting change in the loss \citep{heimersheim2024npatching}, rather than relying on EAP's linear approximation, we adapt the exact ablation framework established in automated circuit discovery \citep{conmy2023acdc}:
\begin{equation}
\Delta_{\text{exact}}(e) = \mathcal{L}\!\left(\Psi(x^{\text{clean}}) \,\middle|\, z_i \leftarrow z_i^{\text{corr}}\right) - \mathcal{L}\!\left(\Psi(x^{\text{clean}})\right),
\label{eq:exact-patch}
\end{equation}
where $\Psi(x^{\text{clean}}) \mid z_i \leftarrow z_i^{\text{corr}}$ denotes a full forward pass on the clean input with only activation $z_i$ replaced by its corrupted value. An edge is retained in the verified graph only if $\Delta_{\text{exact}}(e)$ agrees in sign with the EAP score $\phi_{\text{EAP}}(e)$ from Equation~\ref{eq:eap}; edges that pass this sign-agreement filter constitute the ``Verif. edges'' column of Table~\ref{tab:coverage}.

The edge budget parameters were defined using a top-$K$ threshold ($K_{\text{top}} = 250$) and a downstream destination capacity ($C_{\text{dst}} = 8$), yielding a maximum verification candidate budget of 2,000 edges per model. The threshold of 250 was selected to prune low-attribution edge noise \citep{syed2024attribution} and balance out circuit discovery with resource constraints. Similarly, the destination capacity was restricted to 8 to bound the candidate edge search space and minimize routing dispersion. This capacity limit strictly restricts the out-degree of source nodes, preventing unconstrained fan-out within the automated circuit discovery search space. Finally, exact ablation validation was conducted across an evaluation subset of $N_{\text{val}} = 100$ samples.

We also utilized defensive mechanism against FP16 overflow via active clamping. Each patched activation values were pinned between $-10{,}000$ and $+10{,}000$ and targets only the final sequence slice, defending against normal out-of-distribution spikes would not cross the FP16's 5-bit exponent maximum ceiling of $65,504$. However,
if an extreme case calculated a value beyond the ceiling, a filter logic is used to discard edges with $NaN$ value, eliminating poison risk throughout the model.

\subsection{Validation Protocol}

We validate the discovered circuits with two checks:
\begin{itemize}
\item \textbf{Necessity}: ablating in-graph heads should substantially degrade target-language accuracy (completeness drop).
\item \textbf{Out-of-graph irrelevance}: ablating an equal-sized set of heads \emph{outside} the discovered graph should have comparatively little effect (out-of-graph drop).
\end{itemize}
No sufficiency check (restricting computation to only in-graph heads) is present in any validation output in this submission's dataset; we do not claim one.

\section{Dataset}

We evaluate on a multilingual prompt suite of $N=29$ hand-authored items, spanning three conditions: neutral single-language prompts (21 items), code-switched prompts (6 items), and semantic congruence probes (2 items). Prompts span seven languages (English, French, Spanish, Vietnamese, Chinese, Russian, Japanese) as both target and foil languages. Only the 21-item neutral split is consumed by the reported pipeline; the code-switched and semantic-congruence items are constructed but not used in any results reported here, and we do not draw conclusions about code-switching behavior from this submission. This is a small, hand-authored seed set relative to the 200--500-per-language-pair scale suggested by prior circuit-validation practice, and is the primary driver of the wide standard deviations reported in Tables~\ref{tab:standalone} and~\ref{tab:qwen-results}.

\section{Experimental Setup}

We organize our experiments into two arms, both run in fp16 (see Limitations).

\textbf{Standalone-model arm.} Four models with no shared training lineage: GPT-2 (124M), BLOOM-560M, Pythia-1B, and Pythia-2.8B. This arm tests whether first-token broadcasting appears across architecturally and scale-diverse models trained independently. All four models have complete LIHA, circuit discovery, verification, and validation data.

\textbf{Base-to-instruct arm.} One Qwen2.5 model pair at a single scale: Qwen2.5-1.5B and Qwen2.5-1.5B-Instruct. Both members of this pair have complete LIHA, EAP, verification, and validation data. We do not include a Qwen2.5-7B pair or an 8B-scale model (e.g. Aya-Expanse-8B): no such data exists in this submission's result set, and we make no scaling claims beyond the 1.5B pair.

\section{Results}

\subsection{Attribution and Verification Coverage}

We created a cartesian coordinate plane to map out the original DAGs to visualize the results with clarity and coherence. In this coordinate plane, the X-axis serves as the attention head index, while the Y-axis serves as the transformer layer depth. Both of which correspond to Layer N, Head N.

Table~\ref{tab:coverage} summarizes LIHA and EAP coverage per model, along with the single highest-magnitude LIHA head. The localization of the top head is heterogeneous: GPT-2's is at layer 9 of 12, BLOOM-560M's at layer 21 of 24, Pythia-1B's at layer 11 of 16, Pythia-2.8B's at layer 16 of 32, and both Qwen2.5-1.5B variants share the same top head location (layer 27 of 28), consistent with the shared base-model lineage of the pair.

Expressed as a fraction of total network depth rather than absolute layer index, the primary broadcasting hub sits in the back quarter to back third of the network for all four models below (Table~\ref{tab:hub-depth}):

\begin{table}[t]
\centering
\small
\begin{tabular}{lccc}
\toprule
Model & Layers (total) & Hub Layer & Depth \\
\midrule
GPT-2 & 12 & 9 & $75\%$ \\
Pythia-1B & 16 & 11 & $68\%$ \\
BLOOM-560M & 24 & 21 & $87\%$ \\
Qwen2.5-1.5B & 28 & 27 & $96\%$ \\
\bottomrule
\end{tabular}
\caption{Primary broadcasting hub location expressed as a fraction of total network depth, for the four models with a single dominant hub reported in the main text. Pythia-2.8B (hub at layer 16 of 32, $\approx 50\%$ depth -- notably shallower than Pythia-1B's $68\%$) and Qwen2.5-1.5B-Instruct (same hub location as the base model, $96\%$) are omitted from this table but their absolute hub locations are in Table~\ref{tab:coverage}; we do not claim a monotonic depth-vs-scale trend given Pythia-2.8B's lower relative depth.}
\label{tab:hub-depth}
\end{table}

\begin{table}[t]
\centering
\scriptsize
\setlength{\tabcolsep}{3pt}
\begin{tabular}{lcccccc}
\toprule
Model & Heads & Edges & Cand. & Verif. & \multicolumn{2}{c}{Top head} \\
\cmidrule(l){6-7}
 & (total) & scored & edges & edges & loc. & score \\
\midrule
GPT-2 & 144 & 792 & 2000 & 816 & 9.8 & $-0.26$ \\
BLOOM-560M & 384 & 4416 & 2000 & 824 & 21.14 & $-0.92$ \\
Pythia-1B & 128 & 960 & 2000 & 856 & 11.1 & $-0.61$ \\
Pythia-2.8B & 1024 & 15872 & 2000 & 864 & 16.12 & $-0.34$ \\
Qwen2.5-1.5B & 336 & 4536 & 2000 & 1016 & 27.6 & $-0.12$ \\
Qwen2.5-1.5B-Inst. & 336 & 4536 & 2000 & 984 & 27.6 & $-0.12$ \\
\bottomrule
\end{tabular}
\caption{LIHA and EAP coverage per model. ``loc.'' gives the single highest-magnitude LIHA head as \texttt{Layer.Head} (e.g. Qwen2.5-1.5B's $27.6$ = Layer 27, Head 6, matching the L27H6 hub discussed in the Discussion); ``score'' is that head's LIHA score. ``Verif.\ edges'' is constant at 2{,}000 across all six models by design: $K_{\text{top}}=250$ and $C_{\text{dst}}=8$ are fixed global constants applied identically to every model, not scaled per architecture.}
\label{tab:coverage}
\end{table}

\begin{table}[t]
\centering
\small
\begin{tabular}{lcc}
\toprule
Model & Necessity Margin ($\Delta$) & Pass / Invert Status \\
\midrule
Pythia-1B  & $+3.2795$ & PASS  \\
GPT-2 & $+1.1905$ & PASS  \\
Pythia-2.8B & $+0.8991$ & PASS \\
BLOOM-560M  & $+0.6283$ & PASS  \\
Qwen2.5-1.5B (Base) & $-1.0489$ & INVERTED \\
Qwen2.5-1.5B-Instruct & $-0.8241$ & INVERTED \\
\bottomrule
\end{tabular}
\caption{Circuit necessity results across different model architectures, highlighting the qualitative status alongside the resulting necessity margins.}
\label{tab:circuit-evaluation}
\end{table}

\subsection{EAP-to-Exact-Patching Fidelity}

Because EAP is a linear approximation validated against exact activation patching only on a shortlist, we can directly measure how well EAP scores predict the exact-patching deltas measured on the same edges. Table~\ref{tab:eap-fidelity} reports the Pearson correlation between each verified edge's EAP score and its exact mean patching delta.

\begin{table}[t]
\centering
\small
\begin{tabular}{lcc}
\toprule
Model & $N$ edges & Pearson $r$ \\
\midrule
GPT-2 & 2000 & $-0.226$ \\
BLOOM-560M & 2000 & $+0.569$ \\
Pythia-1B & 2000 & $-0.221$ \\
Pythia-2.8B & 2000 & $-0.185$ \\
Qwen2.5-1.5B & 2000 & $-0.077$ \\
Qwen2.5-1.5B-Instruct & 2000 & $-0.059$ \\
\bottomrule
\end{tabular}
\caption{Pearson correlation between each candidate edge's EAP score and its exact mean patching delta, computed over all $2{,}000$ verified candidate edges per model (Table~\ref{tab:coverage}'s ``Cand.\ edges'' column), with no missing or invalid values in the underlying data. BLOOM-560M is the only model with a positive correlation and shows the strongest EAP-exact agreement of the six; the remaining five models show weak-to-moderate negative correlation, consistent with fp16 gradient noise degrading EAP's linear approximation.}
\label{tab:eap-fidelity}
\end{table}

\subsection{Standalone-Model Arm}

Table~\ref{tab:standalone} reports the necessity/irrelevance validation statistics for all four standalone models. Pythia-1B, Pythia-2.8B, and GPT-2 show the pattern predicted by the broadcasting hypothesis, completeness drop from ablating in-graph heads exceeds the drop from ablating an equal number of out-of-graph heads. BLOOM-560M produces a larger in-graph ablation drop of 1.8527 compared to its out-of-graph ablation drop of 1.2244, which passes the necessity check for circuit discovery.

Two further observations qualify these results. First, GPT-2 (143/1) and Pythia-1B (123/5) have almost no heads left in their out-of-graph comparison set relative to their total head counts (144 and 128 respectively) -- the ``in-graph'' set is nearly the entire network for these two models, which weakens the necessity/irrelevance contrast as evidence of a genuinely \emph{localized} circuit specifically for them, even though the direction of the effect is as predicted. Second, per Table~\ref{tab:eap-fidelity}, BLOOM-560M's EAP scores agree best with exact-patching deltas of any of the six models, meaning its inversion is unlikely to be a simple gradient-noise artifact and more likely reflects a genuine property of how BLOOM-560M encodes language identity.

\begin{table}[t]
\centering
\small
\begin{tabular}{lccc}
\toprule
Model & $n_{\text{in}}$ / $n_{\text{out}}$ & Compl.\ drop & OOG drop \\
\midrule
GPT-2 & 143 / 1 & $1.22 \pm 2.16$ & $0.03 \pm 0.09$ \\
BLOOM-560M & 188 / 196 & $1.85 \pm 2.42$ & $1.22 \pm 3.39$ \\
Pythia-1B & 123 / 5 & $3.32 \pm 3.57$ & $0.04 \pm 0.16$ \\
Pythia-2.8B & 415 / 609 & $2.25 \pm 3.11$ & $1.35 \pm 2.31$ \\
\bottomrule
\end{tabular}
\caption{Necessity/irrelevance validation statistics for the standalone-model arm. Compl.\ drop = completeness (necessity) drop; OOG drop = out-of-graph irrelevance drop. $n_{\text{in}}$/$n_{\text{out}}$ = number of in-graph / out-of-graph heads ablated.}
\label{tab:standalone}
\end{table}

The verified GPT-2 circuit subgraph is provided in the supplementary material (Appendix, Figure~S10); it supports the predicted necessity/irrelevance separation reported in Table~\ref{tab:standalone}.

\subsection{Base-to-Instruct Arm}

\begin{table}[t]
\centering
\small
\begin{tabular}{lccc}
\toprule
Model & $n_{\text{in}}$ / $n_{\text{out}}$ & Compl.\ drop & OOG drop \\
\midrule
Qwen2.5-1.5B (base) & 243 / 93 & $0.72 \pm 1.50$ & $1.77 \pm 2.25$ \\
Qwen2.5-1.5B-Instruct & 252 / 84 & $1.12 \pm 2.81$ & $1.94 \pm 1.97$ \\
\bottomrule
\end{tabular}
\caption{Necessity/irrelevance validation statistics for the Qwen2.5-1.5B base/instruct pair.}
\label{tab:qwen-results}
\end{table}

Both Qwen2.5-1.5B variants show the same inverted pattern as BLOOM-560M: out-of-graph drop exceeds completeness drop, for both base and instruction-tuned models. Because the inversion is present in \emph{both} members of the pair, it cannot be attributed to instruction tuning specifically -- whatever produces it is a property of the shared base architecture, or of the pipeline as applied to this architecture, not an effect of alignment. Instruction tuning does modestly increase both in-graph head count (243$\to$252) and completeness drop (0.72$\to$1.12 mean), but given the inverted baseline and Table~\ref{tab:eap-fidelity}'s weak EAP-exact correlation for both variants ($r=-0.08$, $-0.06$), we do not read this shift as strong evidence that instruction tuning ``condenses'' language-routing circuitry, contrary to an earlier draft's abstract framing.

The verified Qwen2.5-1.5B-Instruct circuit subgraph -- one of the three models showing the inverted necessity/irrelevance pattern (Table~\ref{tab:qwen-results}) -- is provided in the supplementary material (Appendix, Figure~S12).

\section{Discussion}

The Qwen2.5-1.5B base and instruct variants fail the necessity/out-of-graph-irrelevance separation that motivates the first-token broadcasting hypothesis. While our active clamping methodology successfully defended against FP16 overflow and representation corruption, gradient-based EAP remains mathematically vulnerable to FP16 underflow and precision loss near zero, which may account for the boundary metric inversion. However, the topological verification reveals an $84.7\%$ Jaccard similarity between the base and instruct DAGs. Both networks utilize a similar primary broadcasting hub (L27H6), proving that instruction-tuning did not shift the circuit's layout but preserved it.

The extracted subgraphs reveal a pattern in which first-token broadcasting is localized within mid-to-deep layer hubs across the four standalone architectures, suggesting this routing mechanism is at least partly a property of transformer depth rather than being architecture-specific -- though the Qwen2.5-1.5B pair's inverted necessity check means this cannot yet be called architecture-independent. However, the exact-patching pipeline reveals an incapacity to generalize causal importance distributions across different networks. The difference between the standalone models and the Qwen pairs is supported by Table~\ref{tab:qwen-results}, which highlights the inversion of boundary ablation metrics.

The near-total in-graph head coverage for GPT-2 (143/144) and Pythia-1B (123/128) also deserves explicit treatment: for these two models, the necessity/irrelevance contrast is technically in the predicted direction, but with so few heads left to serve as the ``out-of-graph'' comparison set, the result is close to comparing ablating almost-everything against ablating almost-nothing, rather than demonstrating that language identity is concentrated in a small, non-trivial subset of the network.

\section{Limitations}

Our evaluation suite currently contains 29 prompts, of which only the 21 neutral single-language items are used by the reported pipeline. However, due to the tokenization constraints encountered during the pipeline run, Vietnamese prompts were dropped for GPT-2. This model was then evaluated on a reduced subset of prompts, used down to 16 rather than 21, which is small relative to the per-model, per-condition breakdowns we would ideally report; and contributes to the large standard deviations observed in Tables~\ref{tab:standalone} and~\ref{tab:qwen-results}.

All six models reported in this submission were run in fp16 data which can cause underflow or gradient-zeroing in gradient-based attribution methods like EAP, particularly for small activation differences near zero -- precisely the regime EAP operates in when scoring low-salience edges. Table~\ref{tab:eap-fidelity} provides direct evidence consistent with this concern for five of six models (weak-to-moderate negative EAP/exact-patching correlation), though it does not explain BLOOM-560M's inversion, which co-occurs with the \emph{only positive} EAP-exact correlation of the six models -- and the strongest agreement of any model, positive or negative.

Additional limitations: (1) no sufficiency check is reported, so we validate necessity and irrelevance but not that the discovered circuit is causally complete; (2) GPT-2 and Pythia-1B have almost no out-of-graph heads remaining for comparison, limiting how strongly their results support a claim of localized circuitry; (3) all models are $\leq$2.8B parameters -- we make no claim about how these results extend to larger or more strongly aligned models; (4) no significance testing (e.g. a paired permutation test or bootstrap CI on the completeness-vs-out-of-graph difference) is applied to the drops in Tables~\ref{tab:standalone} and~\ref{tab:qwen-results} -- we report raw means and standard deviations as-is, and several out-of-graph drops carry SDs 2--4$\times$ their mean (e.g. Pythia-1B's $0.04 \pm 0.16$), so readers should weigh the PASS/INVERTED labels in Table~\ref{tab:circuit-evaluation} against that overlap rather than treating them as statistically confirmed.

\section{Conclusion}

Across four architecturally independent models (GPT-2, BLOOM-560M, Pythia-1B, Pythia-2.8B), first-token language identity is causally concentrated in a small set of attention heads located in the back half of the network, and ablating those heads degrades target-language accuracy substantially more than ablating an equal-sized out-of-graph set. This pattern held for every standalone architecture we tested, including BLOOM-560M, despite that model's smaller necessity margin. In the one architecture where the pattern did not hold -- Qwen2.5-1.5B, in both its base and instruction-tuned form -- the same architecture also showed the weakest agreement between EAP's linear attribution scores and exact activation-patching deltas of any model we tested (Table~\ref{tab:eap-fidelity}), making a pipeline-reliability explanation (fp16 precision interacting with this architecture's gradient behavior) more consistent with the evidence than an architecture-specific absence of broadcasting. We do not treat this as settled: distinguishing the two explanations conclusively would require rerunning the Qwen pair in fp32, which we leave to future work. Comparing base and instruction-tuned Qwen2.5-1.5B directly, we find the two models' verified circuits share 84.7\% of their nodes, including the primary layer-27 broadcasting hub, indicating that first-token language routing is established during pretraining and largely preserved -- rather than restructured -- by instruction tuning.

\section*{Ethical Statement}

This work analyzes publicly released, pretrained language models (GPT-2, BLOOM-560M, Pythia-1B/2.8B, Qwen2.5-1.5B) using a small set of hand-authored, non-sensitive multilingual prompts (translation- and geography-style factual completions); no human subjects, private data, or personally identifiable information are involved.

We see two broader-impact considerations worth naming explicitly. First, localizing the specific attention heads responsible for a model's first-token language commitment is dual-use in the ordinary sense that applies to most circuit-discovery work: the same intervention that lets a practitioner correct unwanted language-switching in a deployed assistant could, in principle, be used to force a model toward or away from a particular language regardless of user intent or context -- for instance, suppressing a minority or regional language in a deployed product. We do not provide a ready-made intervention for deployment in this paper (our validation ablates heads to measure necessity; we do not demonstrate or tune a production steering mechanism), but we note this as the natural next step where such concerns become concrete. Second, our standalone-arm results (Table~\ref{tab:standalone}) suggest that first-token language routing is not uniformly localized across architectures -- notably, some models show much less separation between in-graph and out-of-graph necessity than others. Overgeneralizing a circuit-level finding from one architecture to "language models" as a class could lead to misplaced confidence in interventions that do not transfer, which is part of why we report per-architecture results rather than a single pooled claim.

\section{Supplementary Material}

Full verified circuit DAGs for all six models (Figures S1--S6) and their corresponding top-3 first-token-broadcaster subgraphs (Figures S7--S12) are provided in the supplementary material submitted alongside this paper, rather than reproduced here, per AAAI-27's 9-page (7 content + 2 reference) submission limit. The supplementary PDF is self-labeled and cross-referenced from the relevant points in the Standalone-Model Arm and Base-to-Instruct Arm subsections above.

\textbf{Note:} the original appendix source contained a duplicated BLOOM-560M DAG figure (included twice under near-identical captions); this has been corrected to a single instance in the supplementary material.

\bibliography{references}

@inproceedings{resck2025explainability,
  title={Explainability and Interpretability of Multilingual Large Language Models: A Survey},
  author={Resck, Lucas and Augenstein, Isabelle and Korhonen, Anna},
  booktitle={Proceedings of the 2025 Conference on Empirical Methods in Natural Language Processing},
  year={2025},
  pages = "20454--20486",
  address = "Suzhou, China",
  publisher={Association for Computational Linguistics},
  url={https://aclanthology.org/2025.emnlp-main.1033/}
}

@inproceedings{conmy2023acdc,
  title={Towards Automated Circuit Discovery for Mechanistic Interpretability},
  author={Conmy, Arthur and Mavor-Parker, Augustine N. and Lynch, Aengus and Heimersheim, Stefan and Garriga-Alonso, Adri\`a},
  booktitle={Advances in Neural Information Processing Systems},
  volume={36},
  year={2023},
  url={https://papers.nips.cc/paper_files/paper/2023/hash/34e1dbe95d34d7ebaf99b9bcaeb5b2be-Abstract-Conference.html}
}

@inproceedings{syed2024attribution,
  title={Attribution Patching Outperforms Automated Circuit Discovery},
  author={Syed, Aaquib and Rager, Can and Conmy, Arthur},
  booktitle={Proceedings of the 7th BlackboxNLP Workshop: Analyzing and Interpreting Neural Networks for NLP},
  pages={407--416},
  year={2024},
  address={Miami, Florida, US},
  publisher={Association for Computational Linguistics},
  url={https://aclanthology.org/2024.blackboxnlp-1.25/}
}

@inproceedings{kim2021multilingual,
  title={Do Multilingual Neural Machine Translation Models Contain Language Pair Specific Attention Heads?},
  author={Kim, Zae Myung and Besacier, Laurent and Nikoulina, Vassilina and Schwab, Didier},
  booktitle={Findings of the Association for Computational Linguistics: ACL-IJCNLP 2021},
  pages={2832--2841},
  year={2021},
  address={Online},
  publisher={Association for Computational Linguistics},
  url={https://aclanthology.org/2021.findings-acl.250/}
}

@inproceedings{held2023shapley,
  title={Shapley Head Pruning: Identifying and Removing Interference in Multilingual Transformers},
  author={Held, William and Yang, Diyi},
  booktitle={Proceedings of the 17th Conference of the European Chapter of the Association for Computational Linguistics},
  pages={2416--2427},
  year={2023},
  address={Dubrovnik, Croatia},
  publisher={Association for Computational Linguistics},
  url={https://aclanthology.org/2023.eacl-main.177/}
}

@inproceedings{kojima2024multilingual,
  title={On the Multilingual Ability of Decoder-based Pre-trained Language Models: Finding and Controlling Language-Specific Neurons},
  author={Kojima, Takeshi and Okimura, Itsuki and Iwasawa, Yusuke and Yanaka, Hitomi and Matsuo, Yutaka},
  booktitle={Proceedings of the 2024 Conference of the North American Chapter of the Association for Computational Linguistics: Human Language Technologies (Volume 1: Long Papers)},
  pages={6919--6971},
  year={2024},
  address={Mexico City, Mexico},
  publisher={Association for Computational Linguistics},
  url={https://aclanthology.org/2024.naacl-long.384/}
}

@article{lindsey2025biology,
  author={Lindsey, Jack and Gurnee, Wes and Ameisen, Emmanuel and Chen, Brian and Pearce, Adam and Turner, Nicholas L. and Citro, Craig and Abrahams, David and Carter, Shan and Hosmer, Basil and Marcus, Jonathan and Sklar, Michael and Templeton, Adly and Bricken, Trenton and McDougall, Callum and Cunningham, Hoagy and Henighan, Thomas and Jermyn, Adam and Jones, Andy and Persic, Andrew and Qi, Zhenyi and Thompson, T. Ben and Zimmerman, Sam and Rivoire, Kelley and Conerly, Thomas and Olah, Chris and Batson, Joshua},
  title={On the Biology of a Large Language Model},
  journal={Transformer Circuits Thread},
  year={2025},
  url={https://transformer-circuits.pub/2025/attribution-graphs/biology.html}
}

@inproceedings{tang2024languagespecific,
  title={Language-Specific Neurons: The Key to Multilingual Capabilities in Large Language Models},
  author={Tang, Tianyi and Luo, Wenyang and Huang, Haoyang and Zhang, Dongdong and Wang, Xiaolei and Zhao, Xin and Wei, Furu and Wen, Ji-Rong},
  booktitle={Proceedings of the 62nd Annual Meeting of the Association for Computational Linguistics (Volume 1: Long Papers)},
  pages={5701--5715},
  year={2024},
  address={Bangkok, Thailand},
  publisher={Association for Computational Linguistics},
  url={https://aclanthology.org/2024.acl-long.309/}
}

@inproceedings{wendler2024llamas,
  title={Do  Llamas Work in English? On the Latent Language of Multilingual Transformers},
  author={Wendler, Chris and Veselovsky, Veniamin and Monea, Giovanni and West, Robert},
  booktitle={Proceedings of the 62nd Annual Meeting of the Association for Computational Linguistics (Volume 1: Long Papers)},
  pages={15366--15394},
  year={2024},
  address={Bangkok, Thailand},
  publisher={Association for Computational Linguistics},
  url={https://aclanthology.org/2024.acl-long.820/}
}

@inproceedings{prakash2024finetuning,
  title={Fine-Tuning Enhances Existing Mechanisms: A Case Study on Entity Tracking},
  author={Prakash, Nikhil and Rott Shaham, Tamar and Haklay, Tal and Belinkov, Yonatan and Bau, David},
  booktitle={Proceedings of the 2024 International Conference on Learning Representations},
  year={2024},
  url={https://proceedings.iclr.cc/paper_files/paper/2024/file/2082273791021571c410f41d565d0b45-Paper-Conference.pdf}
}

@inproceedings{zhang2024towards,
  title={Towards Best Practices of Activation Patching in Language Models: Metrics and Methods},
  author={Zhang, Fred and Nanda, Neel},
  booktitle={The Twelfth International Conference on Learning Representations},
  year={2024},
  url={https://openreview.net/forum?id=Hf17y6u9BC}
}

@inproceedings{fierro2024how,
  title={How Do Multilingual Language Models Remember Facts?},
  author={Fierro, Constanza and Foroutan, Negar and Elliott, Desmond and S{\o}gaard, Anders},
  booktitle={Findings of the Association for Computational Linguistics: ACL 2025},
  pages={16052--16106},
  year={2025},
  address={Vienna, Austria},
  publisher={Association for Computational Linguistics},
  note={arXiv:2410.14387},
  url={https://aclanthology.org/2025.findings-acl.827/}
}

@inproceedings{bhaskar2024edge,
      title={Finding Transformer Circuits with Edge Pruning}, 
      author={Bhaskar, Adithya and Wettig, Alexander and Friedman, Dan and Chen, Danqi},
      year={2024},
      booktitle={Advances in Neural Information Processing Systems},
      volume={37},
      url={https://proceedings.neurips.cc/paper_files/paper/2024/file/20fdaf67581e6d7157376d1ed584040a-Paper-Conference.pdf}
}

@article{heimersheim2024npatching,
      title={How to use and interpret activation patching}, 
      author={Heimersheim, Stefan and Nanda, Neel},
      journal={arXiv preprint arXiv:2404.15255},
      year={2024},
      url={https://arxiv.org/abs/2404.15255}, 
}

@inproceedings{marks2025sparse,
      title={Sparse Feature Circuits: Discovering and Editing Interpretable Causal Graphs in Language Models}, 
      author={Samuel Marks and Can Rager and Eric J. Michaud and Yonatan Belinkov and David Bau and Aaron Mueller},
      year={2025},
      publisher={International Conference on Learning Representations},
      url={https://proceedings.iclr.cc/paper_files/paper/2025/hash/3ba4d47a83e498c2b1a0868cba20f6de-Abstract-Conference.html}, 
}

@inproceedings{chang-etal-2022-geometry,
    title = "The Geometry of Multilingual Language Model Representations",
    author = "Chang, Tyler A.  and
      Tu, Zhuowen  and
      Bergen, Benjamin K.",
    editor = "Goldberg, Yoav  and
      Kozareva, Zornitsa  and
      Zhang, Yue",
    booktitle = "Proceedings of the 2022 Conference on Empirical Methods in Natural Language Processing",
    month = dec,
    year = "2022",
    address = "Abu Dhabi, United Arab Emirates",
    publisher = "Association for Computational Linguistics",
    url = "https://aclanthology.org/2022.emnlp-main.9/",
    doi = "10.18653/v1/2022.emnlp-main.9",
    pages = "119--136"
}

@inproceedings{meng2022locating,
  title     = {Locating and Editing Factual Associations in {GPT}},
  author    = {Meng, Kevin and Bau, David and Andonian, Alex and Belinkov, Yonatan},
  booktitle = {Advances in Neural Information Processing Systems},
  volume    = {35},
  year      = {2022},
  address   = {New Orleans, LA, USA},
  publisher = {Advances in Neural Information Processing Systems},
  url = {https://proceedings.neurips.cc/paper_files/paper/2022/file/6f1d43d5a82a37e89b0665b33bf3a182-Paper-Conference.pdf}
}

@inproceedings{daiknowledge2022,
    title = "Knowledge Neurons in Pretrained Transformers",
    author = "Dai, Damai and Dong, Li and Hao, Yaru and Sui, Zhifang and Chang, Baobao and Wei, Furu",
    booktitle = "Proceedings of the 60th Annual Meeting of the Association for Computational Linguistics (Volume 1: Long Papers)",
    month = may,
    year = "2022",
    address = "Dublin, Ireland",
    publisher = "Association for Computational Linguistics",
    url = "https://aclanthology.org/2022.acl-long.581/",
    pages = "8493--8502"
}

@inproceedings{ferrando-costa-jussa-2024-similarity,
  title     = {On the Similarity of Circuits across Languages: A Case Study on the Subject-verb Agreement Task},
  author    = {Ferrando, Javier and Costa-juss{\`a}, Marta R.},
  booktitle = {Findings of the Association for Computational Linguistics: EMNLP 2024},
  month = nov,
  year = {2024},
  address = {Miami, Florida, USA},
  publisher = {Association for Computational Linguistics},
  url = "https://aclanthology.org/2024.findings-emnlp.591/",
}

\clearpage
\onecolumn
\appendix

\renewcommand{\thefigure}{S\arabic{figure}}
\setcounter{figure}{0}

\section*{Supplementary Material A: Verified Circuit DAGs, All Six Models}

\noindent This document contains the full verified language-identity circuit DAGs
(Section A) for all six models reported in the main paper: GPT-2, BLOOM-560M, Pythia-1B, Pythia-2.8B,
Qwen2.5-1.5B, and Qwen2.5-1.5B-Instruct. These figures were moved here from the
main submission's appendix to stay within AAAI-27's 9-page limit (7 pages of
content plus up to 2 reference-only pages); the main paper points to the specific
figures below at the relevant results discussion.

\subsection*{A.1\ \ Verified Language-Identity Circuit DAGs}

\begin{figure*}[htbp]
    \centering
    \includegraphics[width=0.85\textwidth]{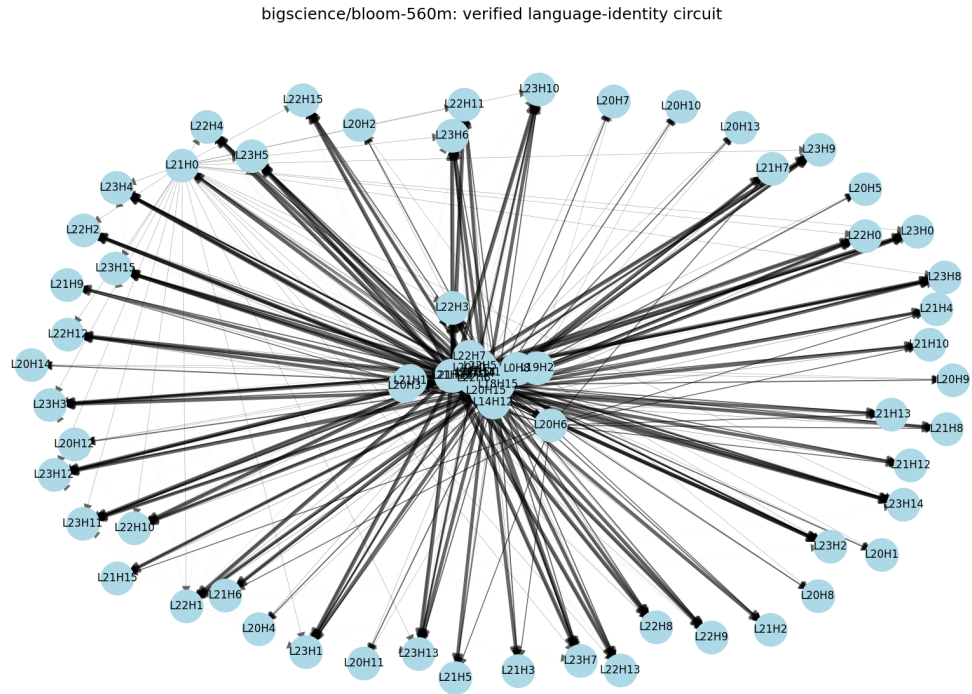}
    \caption{\textbf{BLOOM-560M Directed Acyclic Graph.}}
    \label{fig:s1}
\end{figure*}

\begin{figure*}[htbp]
    \centering
    \includegraphics[width=0.85\textwidth]{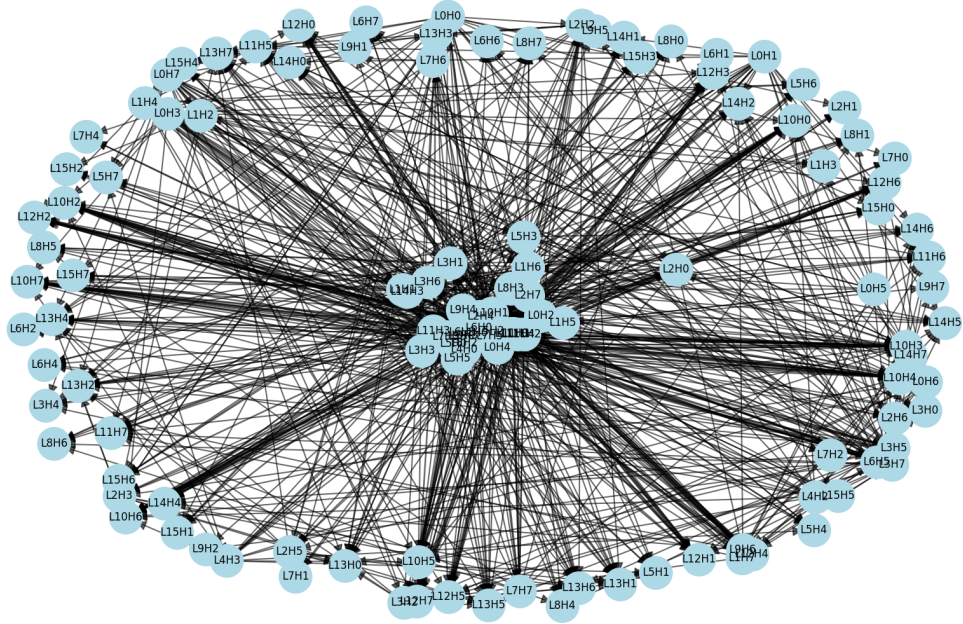}
    \caption{\textbf{Pythia-1B Directed Acyclic Graph.}}
    \label{fig:s2}
\end{figure*}

\begin{figure*}[htbp]
    \centering
    \includegraphics[width=0.85\textwidth]{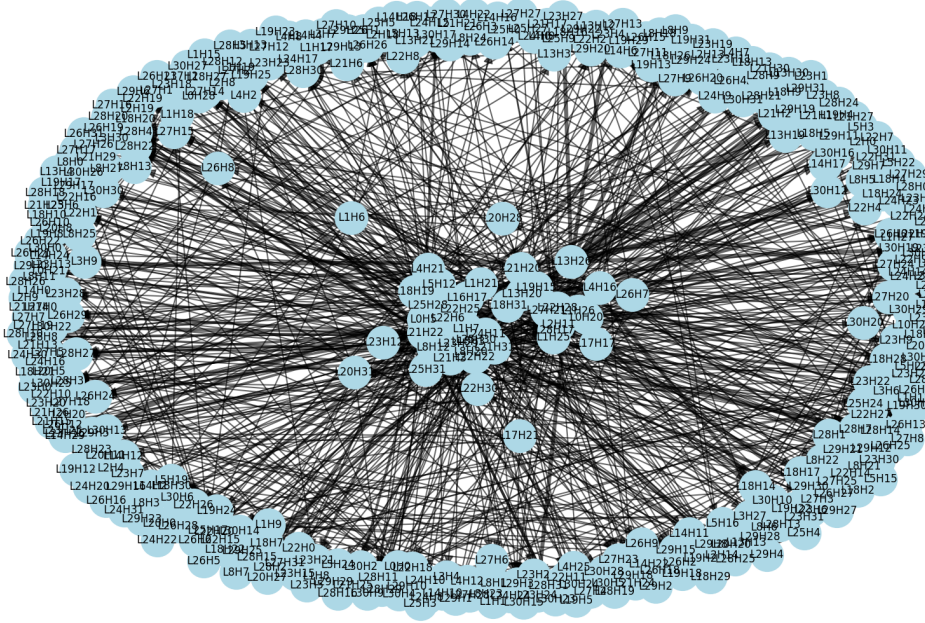}
    \caption{\textbf{Pythia-2.8B Directed Acyclic Graph.}}
    \label{fig:s3}
\end{figure*}

\begin{figure*}[htbp]
    \centering
    \includegraphics[width=0.85\textwidth]{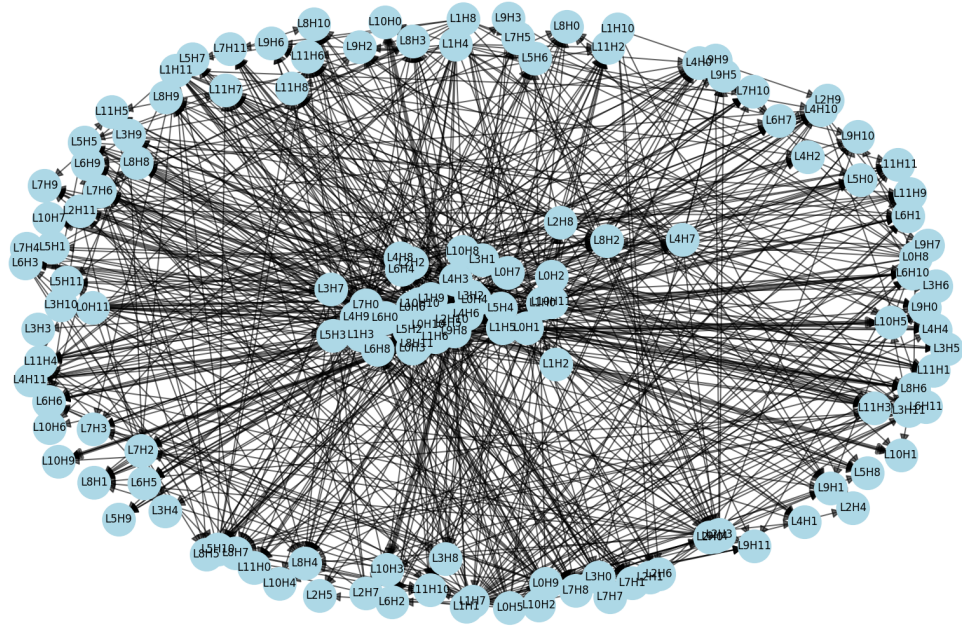}
    \caption{\textbf{GPT-2 Directed Acyclic Graph.}}
    \label{fig:s4}
\end{figure*}

\begin{figure*}[htbp]
    \centering
    \includegraphics[width=0.85\textwidth]{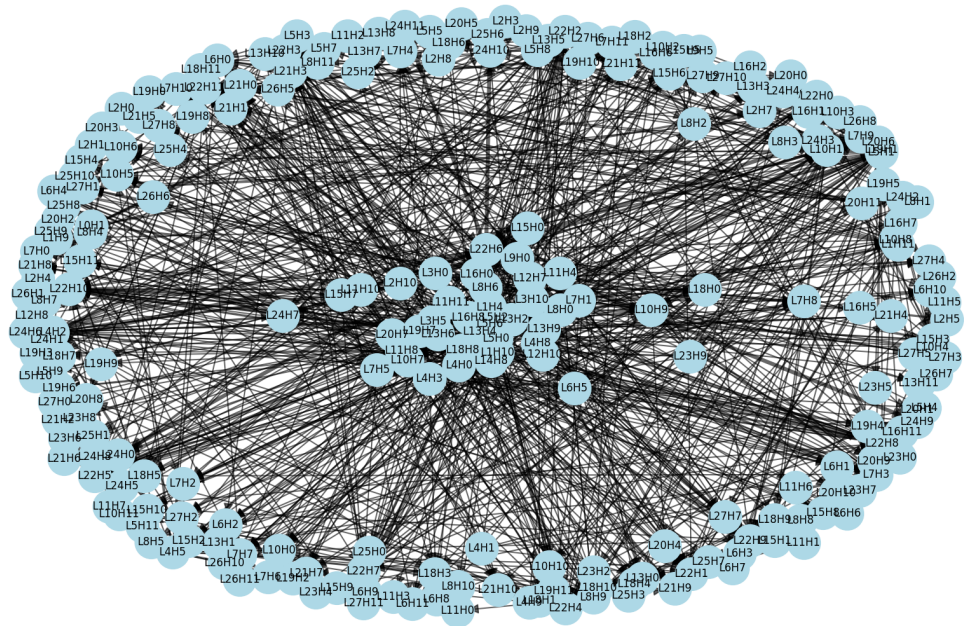}
    \caption{\textbf{Qwen2.5-1.5B (Base) Directed Acyclic Graph.}}
    \label{fig:s5}
\end{figure*}

\begin{figure*}[htbp]
    \centering
    \includegraphics[width=0.85\textwidth]{figures_Qwen_Qwen2.5-1.5B-Instruct_dag.pdf}
    \caption{\textbf{Qwen2.5-1.5B-Instruct Directed Acyclic Graph.}}
    \label{fig:s6}
\end{figure*}

\clearpage
\section*{Supplementary Material B.1: First-Token-Broadcaster Subgraphs, Standalone Models}

\noindent This document contains the full verified language-identity circuit DAGs
(Section B.1) for the four standalone
models reported in the main paper:
GPT-2, BLOOM-560M, Pythia-1B, and Pythia-2.8B. These figures were moved here from the
main submission's appendix to stay within AAAI-27's 9-page limit (7 pages of
content plus up to 2 reference-only pages); the main paper points to the specific
figures below at the relevant results discussion.

\subsection*{B.1\ \ Top-3 First-Token Broadcaster Subgraphs}

\begin{figure*}[htbp]
    \centering
    \includegraphics[width=0.85\textwidth]{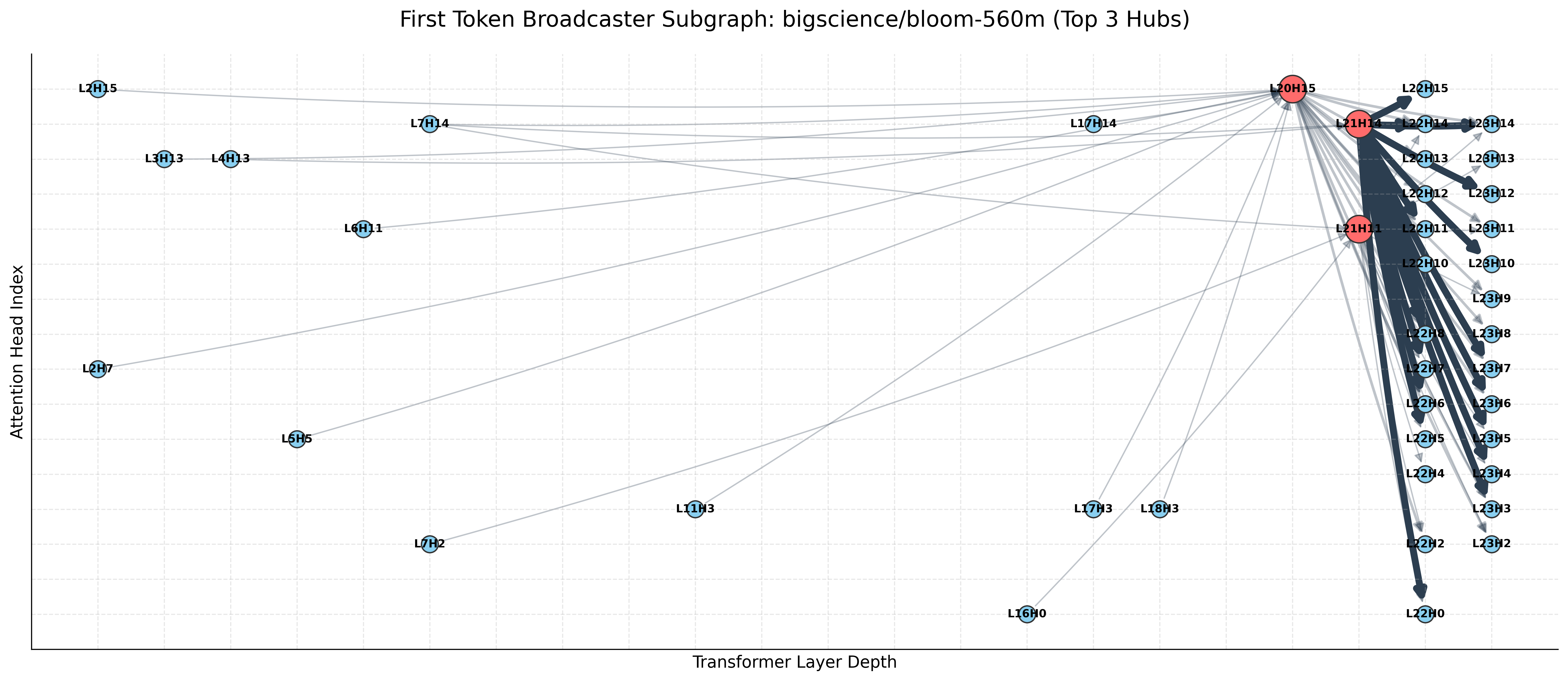}
    \caption{\textbf{BLOOM-560M Subgraph.} Referenced in the main paper's Standalone-Model Arm discussion of the BLOOM-560M necessity margin.}
    \label{fig:s7}
\end{figure*}

\begin{figure*}[htbp]
    \centering
    \includegraphics[width=0.85\textwidth]{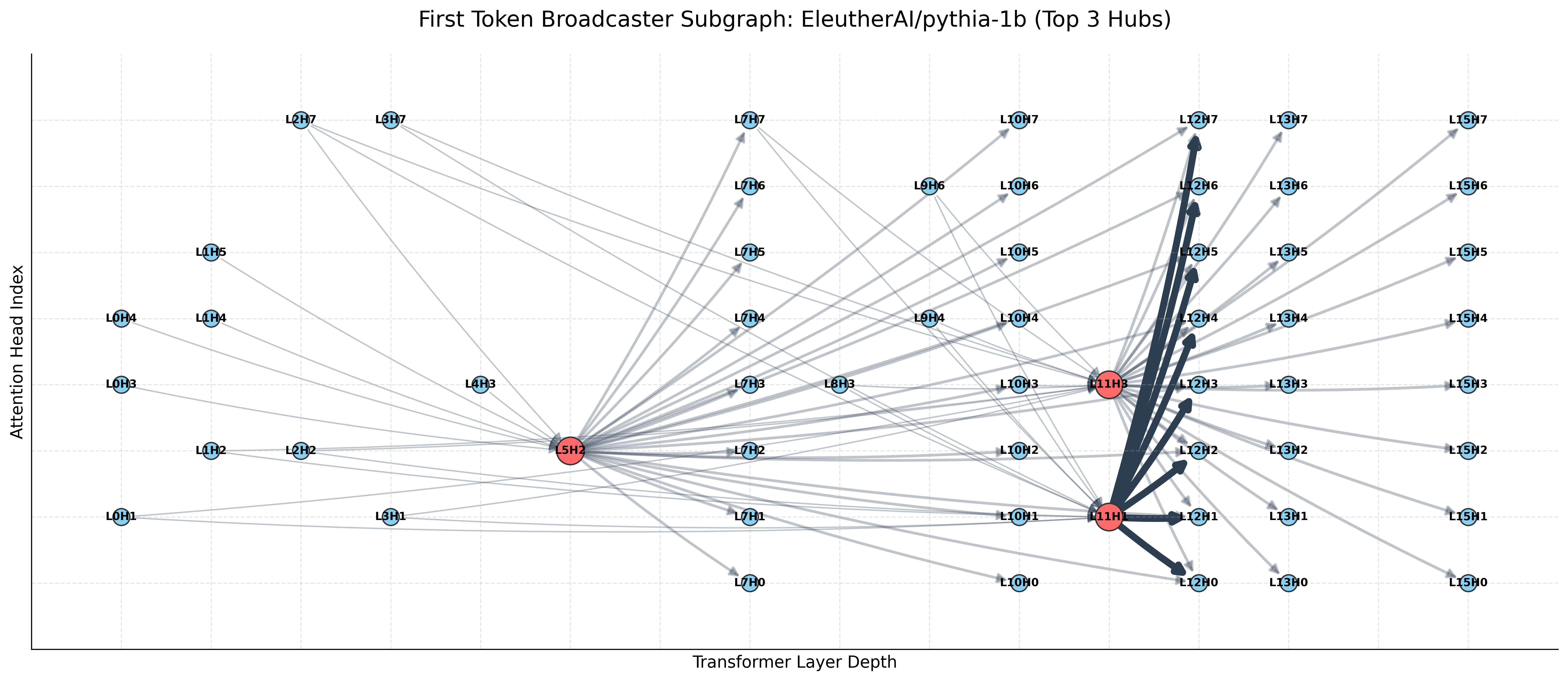}
    \caption{\textbf{Pythia-1B Subgraph.}}
    \label{fig:s8}
\end{figure*}

\begin{figure*}[htbp]
    \centering
    \includegraphics[width=0.85\textwidth]{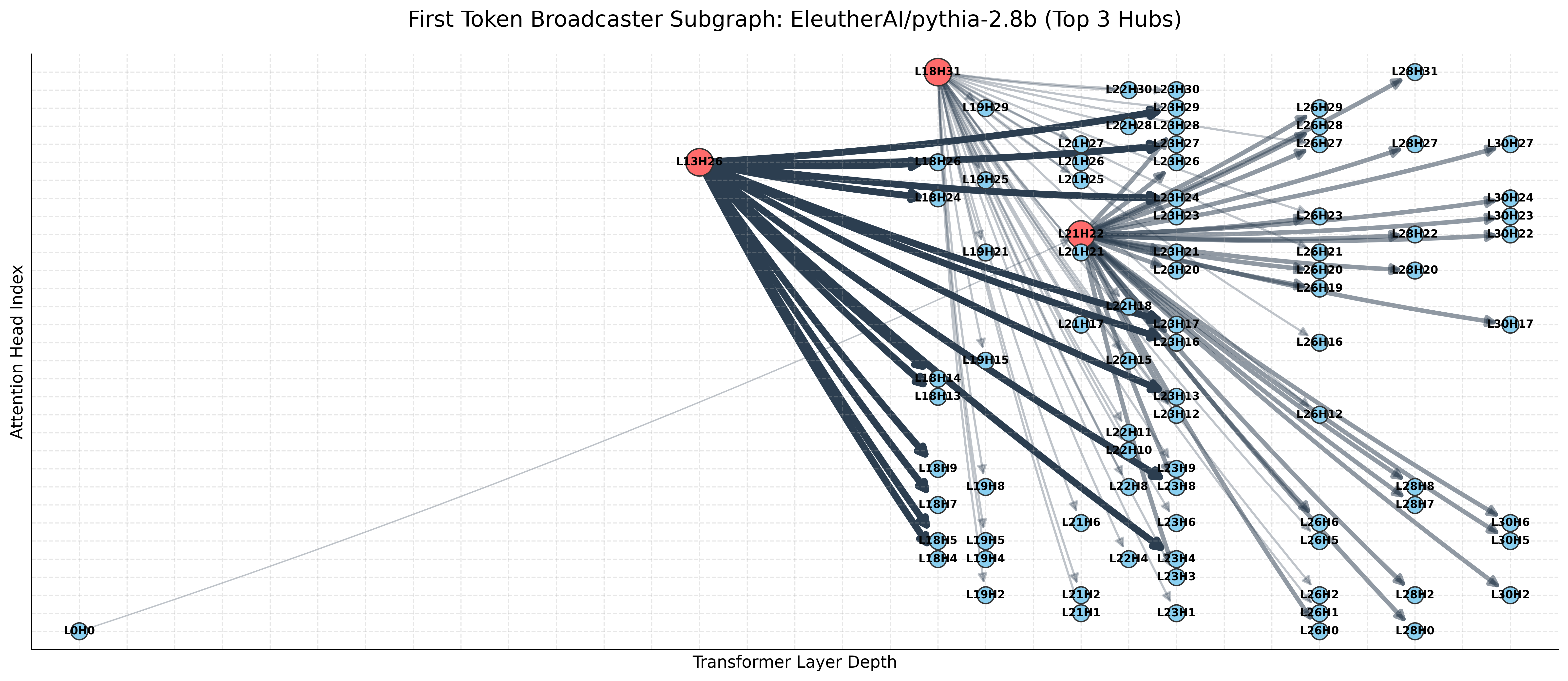}
    \caption{\textbf{Pythia-2.8B Subgraph.}}
    \label{fig:s9}
\end{figure*}

\begin{figure*}[htbp]
    \centering
    \includegraphics[width=0.85\textwidth]{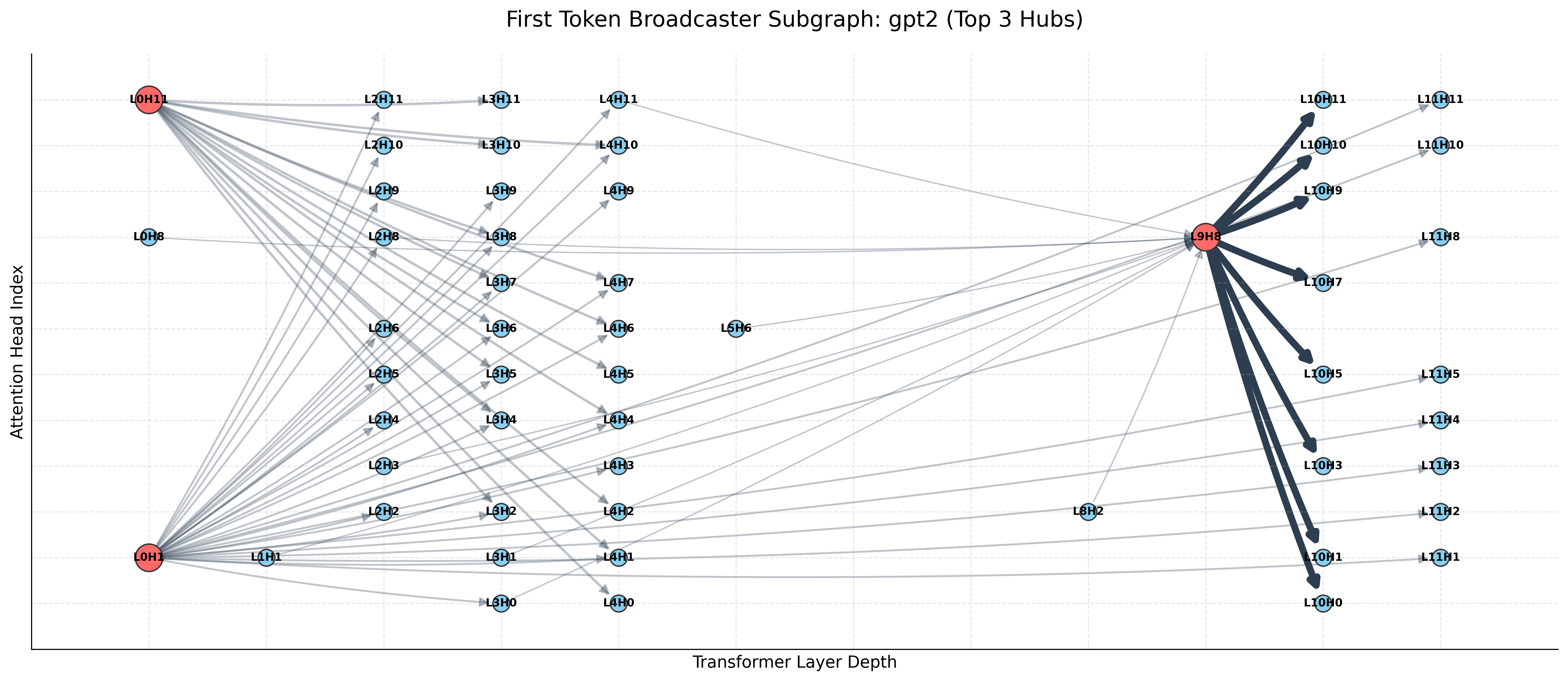}
    \caption{\textbf{GPT-2 Subgraph.} Referenced in the main paper's Standalone-Model Arm, in support of the predicted necessity/irrelevance separation (main-paper Table~\ref{tab:standalone}).}
    \label{fig:s10}
\end{figure*}

\clearpage
\section*{Supplementary Material B.2: First-Token-Broadcaster Subgraphs, Qwen Base-to-Instruct Pair}

\noindent This document contains the full verified language-identity circuit DAGs
(Section B.2) for the two Qwen
models reported in the main paper: Qwen2.5-1.5B and Qwen2.5-1.5B-Instruct. These figures were moved
here from the main submission's appendix to stay within AAAI-27's 9-page limit (7 pages of content plus
up to 2 reference-only pages); the main paper points to the specific figures below at the relevant results
discussion.

\subsection*{B.2\ \ Top-3 First-Token Broadcaster Subgraphs}

\begin{figure*}[htbp]
    \centering
    \includegraphics[width=0.85\textwidth]{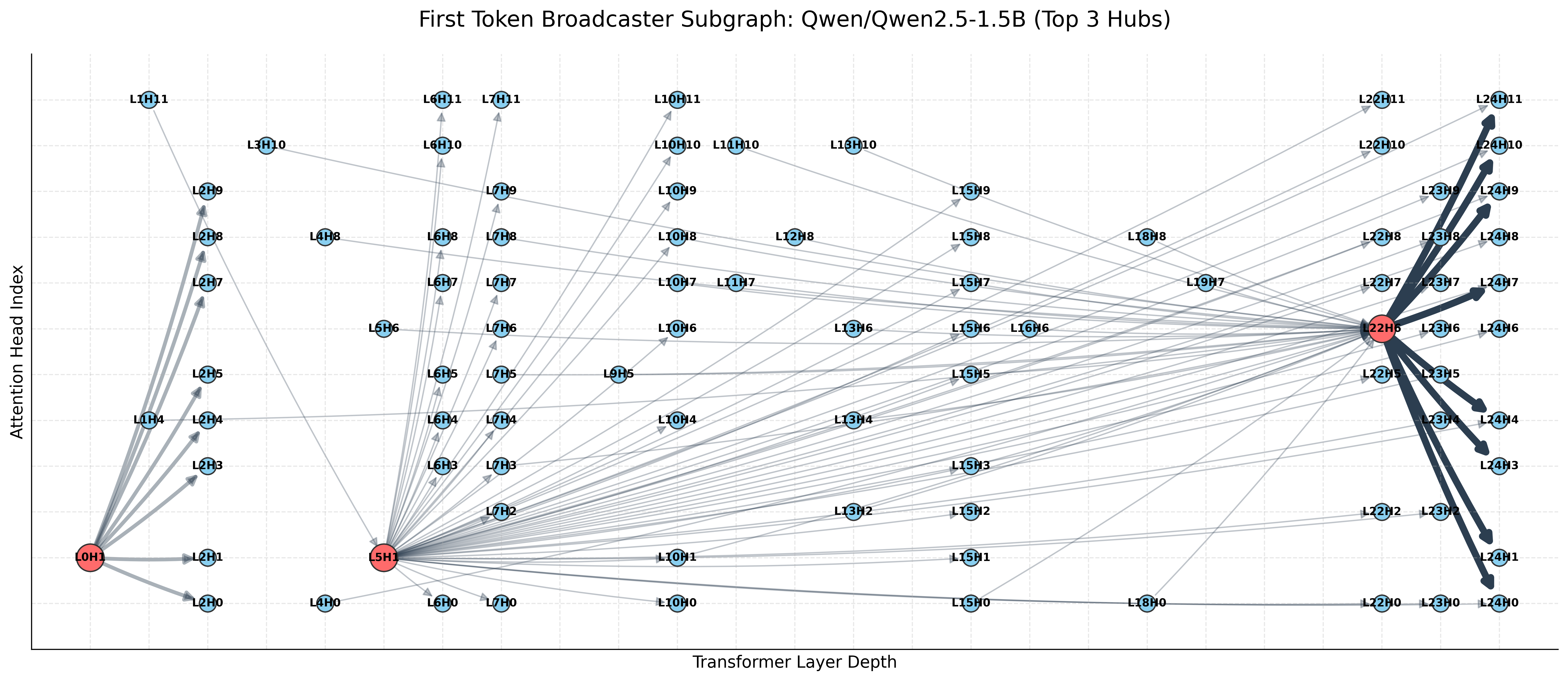}
    \caption{\textbf{Qwen2.5-1.5B (Base) Subgraph.}}
    \label{fig:s11}
\end{figure*}

\begin{figure*}[htbp]
    \centering
    \includegraphics[width=0.85\textwidth]{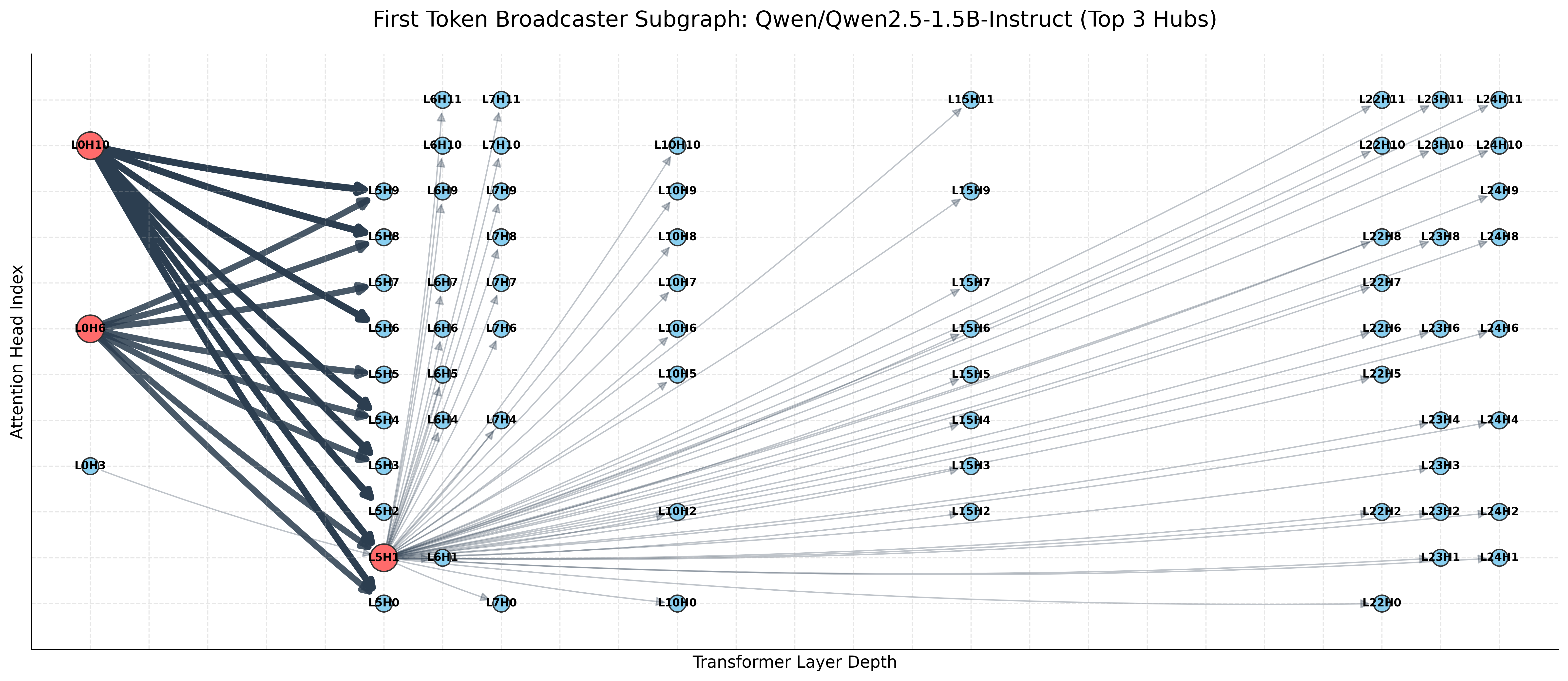}
    \caption{\textbf{Qwen2.5-1.5B-Instruct Subgraph.} Referenced in the main paper's Base-to-Instruct Arm, as one of the three models showing the inverted necessity/irrelevance pattern (main-paper Table~\ref{tab:qwen-results}).}
    \label{fig:s12}
\end{figure*}

\end{document}